\pdfoutput=1

\documentclass[11pt]{article}

\usepackage[]{acl}

\usepackage{times}
\usepackage{latexsym}

\usepackage[T1]{fontenc}

\usepackage[utf8]{inputenc}

\usepackage{microtype}

\usepackage{hyperref}       
\usepackage{url}            
\usepackage{booktabs}       
\usepackage{amsfonts}       
\usepackage{nicefrac}       
\usepackage{graphicx}       
\usepackage{natbib}       
\usepackage{amsmath, amssymb, amsfonts, bm, mathtools}       
\usepackage{courier}

\usepackage{makecell}
\usepackage{multirow}
\usepackage{subfigure}
\usepackage{enumitem}
\usepackage{tabularx}
\usepackage{xspace}
\usepackage{adjustbox}
\usepackage{longtable}
\usepackage{listings}
\usepackage[normalem]{ulem} 

\newcommand{\secref}[2][]{Section#1~\ref{sec:#2}}

\newcommand{\tabref}[2][]{Table#1~\ref{tab:#2}}
\newcommand{\figref}[2][]{Figure#1~\ref{fig:#2}}

\newcommand{\appref}[2][]{Appendix#1~\ref{#2}}

\definecolor{Mulberry}{rgb}{0.77,0.29,0.55}
\definecolor{CadmiumOrange}{rgb}{0.93,0.53, 0.18}
\definecolor{ForestGreen}{rgb}{0.13, 0.55, 0.13}
\definecolor{WildStrawberry}{rgb}{0.5, 0.7, 0.2}

\newcommand{\squishlist}{
   \begin{list}{$\bullet$}
    { \setlength{\itemsep}{0pt}      \setlength{\parsep}{3pt}
      \setlength{\topsep}{3pt}       \setlength{\partopsep}{0pt}
      \setlength{\leftmargin}{1.5em} \setlength{\labelwidth}{1em}
      \setlength{\labelsep}{0.5em} } }

\newcommand{\squishend}{
    \end{list}  }

\title{A Sentiment Consolidation Framework for Meta-Review Generation}

\author{Miao Li$^1$ \quad Jey Han Lau$^1$ \quad Eduard Hovy$^{1, 2}$ \\
        $^1$School of Computing and Information Systems, The University of Melbourne \\
        $^2$Language Technologies Institute, Carnegie Mellon University\\
        \texttt{miao4@student.unimelb.edu.au},\\ \texttt{\{laujh, eduard.hovy\}@unimelb.edu.au}
        }

\begin{document}
\maketitle
\begin{abstract}
Modern natural language generation systems with Large Language Models (LLMs) exhibit the capability to generate a plausible summary of multiple documents; however, it is uncertain if they truly possess the capability of information consolidation to generate summaries, especially on documents with opinionated information. We focus on meta-review generation, a form of sentiment summarisation for the scientific domain. To make scientific sentiment summarization more grounded, we hypothesize that human meta-reviewers follow a three-layer framework of sentiment consolidation to write meta-reviews. Based on the framework, we propose novel prompting methods for LLMs to generate meta-reviews and evaluation metrics to assess the quality of generated meta-reviews. Our framework is validated empirically as we find that prompting LLMs based on the framework --- compared with prompting them with simple instructions --- generates better meta-reviews.\footnote{The code and annotated data are accessible at \url{https://github.com/oaimli/MetaReviewingLogic}.}
\end{abstract}

\section{Introduction}

Notable strides have been made in abstractive text summarization~\citep{summarization_survey_2021} with the advancement of Large Language Models (LLMs)~\citep{llm_survey_wxzhao_2023} over recent years. With even a simple instruction such as ``\textit{tl;dr}'' or ``\textit{please write a summary}'', these models can generate plausible summaries which are found more preferred over those written by humans~\citep{dead_summarization_2023}. 
However, it is uncertain if these models truly possess the ability of information consolidation, especially when summarizing documents that are composed of opinionated information. The models may take shortcuts to generate texts instead of correctly understanding and aggregating information from the source documents~\citep{evaluation_gehrmann_2023} and they may generate abstractive summaries with incorrect overall sentiment. 

\begin{figure}[t]
\centering
\includegraphics[width=0.5\textwidth, trim=110 395 100 150, clip]{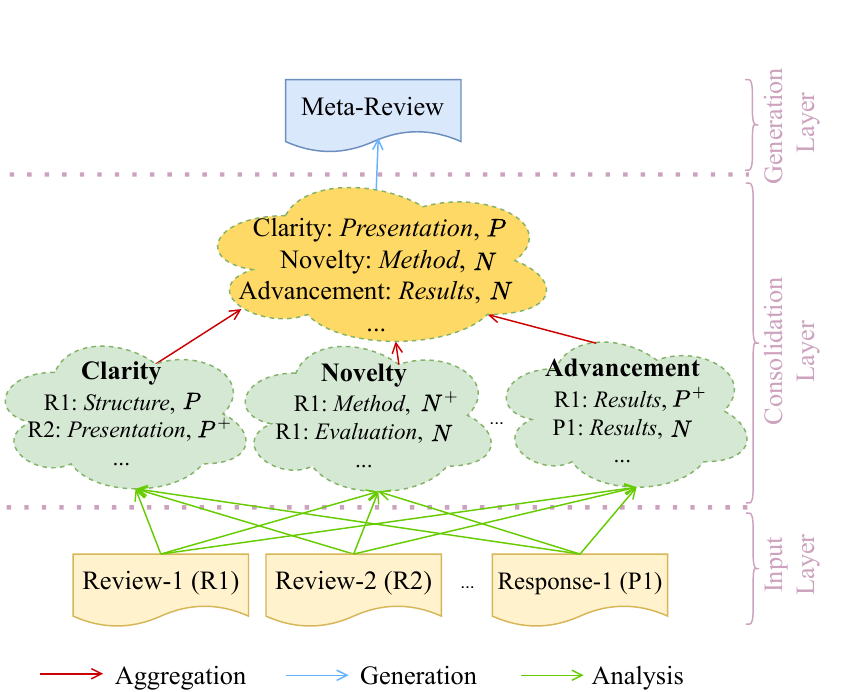}
\caption{The three-layer framework of the underlying information consolidation logic in meta-reviewing ($P$: Positive, $P^+$: Strongly positive, $N$: Negative, $N^+$: Strongly negative). 
}
\label{fig:three-layer-framework}
\end{figure}

Automated sentiment summarization holds significant importance~\citep{kim2011comprehensive} and there have been sentiment summarization datasets; however, most of them are in the product review domain. These datasets are less interesting for investigating information consolidation as (1) the summaries are extractive, composed of a simple combination of extracted snippets~\citep{content_planning_opinion_summarization_2021}, and (2) the summary of product reviews is about extracting the majority sentiment (which is a simple consolidation function). To address this, in this paper, we propose the task of scientific sentiment summarization, taking the meta-reviews in scientific peer review as summaries.\footnote{The representative peer review platform which is publicly available is \url{www.openreview.com}.} The investigation of meta-review generation~\citep{peersum_2023} presents an exciting opportunity for exploring the intricate process of multi-document information consolidation that involves complex judgement. This is because (1) meta-reviewers are supposed to understand not only all the reviews from different reviewers but also the multi-turn discussions between the reviewers and the author and write their comments to support the acceptance decision of the manuscript, (2) the logic of arguments (from reviewers and authors) has to be taken into account to arrive at the final sentiment in the meta-reviews and it is not a matter of majority voting and (3) meta-reviews have to recognize and resolve conflicts and consensus among reviewers.


In this paper, we hypothesize that human meta-reviewers follow a three-layer sentiment consolidation framework as shown in~\figref{three-layer-framework} to write meta-reviews based on reviews and multi-turn discussions in the peer review process. Human and automatic annotation is then conducted to extract sentiments and expressions on various review facets (e.g.,\ novelty and soundness) from corresponding source documents (i.e., reviews and discussions) and these judgements play a critical role in generating the meta-reviews. We also propose two evaluation metrics which focus on assessing sentiments in generated meta-reviews, and experiments empirically validate our proposed three-layer framework when they are integrated as prompts for LLMs to generate meta-reviews.

Contributions of our paper:
\squishlist
    \item We hypothesize that human meta-reviewers follow a three-layer sentiment consolidation framework when writing meta-reviews;
    \item We collect human annotations on meta-reviews and corresponding source documents based on the consolidation framework;
    \item We propose two automatic metrics (reference-free and reference-based) to evaluate the sentiment in the generated meta-reviews. 
    \item Experiments validate the empirical effectiveness of the framework when we incorporate it as prompts for LLMs to generate meta-reviews.
\squishend

\begin{table*}[t]
    \setlength\tabcolsep{3pt}
    \centering
    \begin{tabular}{p{3.5cm}p{8.5cm}}
    \toprule
    \textbf{Component} & \textbf{Definition} \\
    \midrule
    Content Expression & What the sentiment is talking about \\
    Sentiment Expression & The value of the sentiment \\
    Review Facet & The specific review facet that the judgement belongs to \\
    Sentiment Level & The polarity and strength of the sentiment \\
    Convincingness Level & How well the sentiment is justified in the document  \\
    \bottomrule
    \end{tabular}
    \caption{Definitions of components in a judgement.}
    \label{tab:annotation_aspects}
\end{table*}

\begin{table}[t]
    \setlength\tabcolsep{4pt}
    \centering
    \begin{adjustbox}{max width=\linewidth}
    \begin{tabular}{lrrr}
    \toprule
     & \textbf{Min} & \textbf{Max} & \textbf{Average} \\
    \midrule
    \#Documents/Sample & 5 & 30 & 12.4 \\
    \#Words/Sample & 1,541 & 11,901 & 4,260.9 \\
    \#Words/Source document & 10 & 1,562 & 360.5\\
    \#Words/Meta-review & 16 & 648 & 150.9\\
    \bottomrule
    \end{tabular}
    \end{adjustbox}
    \caption{Statistics of the human annotated data.}
    \label{tab:annotation_data_statistics}
\end{table}

\section{Related Work}

In this section, we discuss large-scale information consolidation in abstractive summarization, and automated sentiment summarization.

\subsection{Large-Scale Information Consolidation}

Natural language generation systems are expected to not only have high-quality generations but also have the ability to comprehend the input information, especially for conditional text generation such as multi-document summarization which has to integrate and aggregate information from different source documents~\citep{evaluation_gehrmann_2023}. Most work in the text summarization community only attempts to improve the generation quality of text summarization, such as relevance and faithfulness, without considering the intricate generation process~\citep{pegasusx_2022, summarization_survey_2021, primera_2022}. For example, \citet{hgsum_2023} use heterogeneous graphs to represent source documents and borrow the idea of graph compression to train the summarization model to get improvement of the generated summaries. However, it is uncertain if these models truly possess the ability to consolidate information from different source documents.

\subsection{Automated Sentiment Summarization}
Sentiment summarization aims to summarise the overall sentiment given a set of documents \citep{sentiment_summarization_2004}. However, most datasets for sentiment summarization are in the product review domain~\citep{content_planning_opinion_summarization_2021}, and scientific sentiment summarization is under-explored.
Meta-review generation, which is a typical scenario of scientific sentiment summarization, is to automatically generate meta-reviews based on reviews and the multi-turn discussions between reviewers and the author of the corresponding manuscript~\citep{peersum_2023}. It is mostly modelled as an end-to-end task~\citep{metagen_2020, gbird_2022, mred_2022, dual_view_2020}. Although \citet{peersum_2023} considered the conversational structure of reviews and discussions, their models do not explain how human meta-reviewers write the meta-reviews.
Different from investigating checklist-guided iterative introspection for meta-review generation with prompting~\citep{sos_checklist_2024},
our work is based on a three-layer sentiment consolidation framework and focuses on various review facets, and we explicitly investigate the sentiment fusion process which is arguably an important aspect of meta-review generation.

\section{Sentiment Consolidation Over Multiple Opinionated Documents}

In the following section, we introduce the task of scientific sentiment summarization and our three-layer sentiment consolidation framework in meta-review generation, conduct sentiment and expression extraction, and analyze the fusion process of scientific sentiments.

\subsection{Hierarchical Sentiment Consolidation}

The task is meta-review generation. We use the PeerSum\footnote{\url{https://github.com/oaimli/PeerSum}} dataset where the input is reviews and discussions and the target output is the corresponding human-written meta-review. We should clarify that even though the task is to generate meta-reviews, our focus here is to get the overall sentiment in the meta-reviews to be correct. Our method and evaluation reflect this focus.

Reading the reviewer guidelines from popular academic presses such as ACM and IEEE\footnote{The complete table of official guidelines that we consider is in~\appref{appendix_review_guidelines}.}, we find they are mostly about \textit{judgements} on the quality and merit of the manuscript. These judgements are generally based on six review facets of criteria: \textit{Novelty}, \textit{Soundness}, \textit{Clarity}, \textit{Advancement}, \textit{Compliance} and \textit{Overall quality}. The meta-reviewers must form their final opinion based on these judgements from the reviewers and authors. Looking at the meta-reviewer guidelines for ICLR\footnote{\url{https://iclr.cc/Conferences/2024/SACguide}} and NeurIPS\footnote{\url{https://nips.cc/Conferences/2020/PaperInformation/AC-SACGuidelines}}, it recommends the meta-reviewer to understand and aggregate information from the whole peer-reviewing process. That is, a human meta-reviewer should first identify judgements from reviews and discussions, and then consolidate these opinions from different review facets to write their meta-review.

To conceptualize this, we propose a three-layer framework, as shown in \figref{three-layer-framework}. The three layers include the input layer, the consolidation layer, and the generation layer. The input layer is the input documents of different types: official reviews and multi-turn discussions. The consolidation layer represents how meta-reviewers process the documents: they first identify and extract judgements from different documents, reorganize the judgements based on review facets, and then consolidate the opinions to form the final opinions of each review facet. In the generation layer, the meta-reviewer writes the meta-review to express the final opinions that they have developed from the previous layer.

\subsection{Judgement Identification and Extraction}
\label{sec:judgement_identification_and_extraction}

Judgements lay the foundation of our proposed framework and the whole peer review process. A judgement here expresses sentiment on a review facet and it contains several components: Content Expression, Sentiment Expression, Review Facet, Sentiment Level, and Convincingness Level (definitions are shown in~\tabref{annotation_aspects}, and an example is given in Appendix \figref{annotation-instruction-p2}). To automate judgement identification and extraction,
we first conduct human annotation, and then leverage in-context learning of LLMs to perform more (automatic) annotation.

\begin{figure}[ht]
\centering
\includegraphics[width=0.49\textwidth, trim=0 20 0 0, clip]{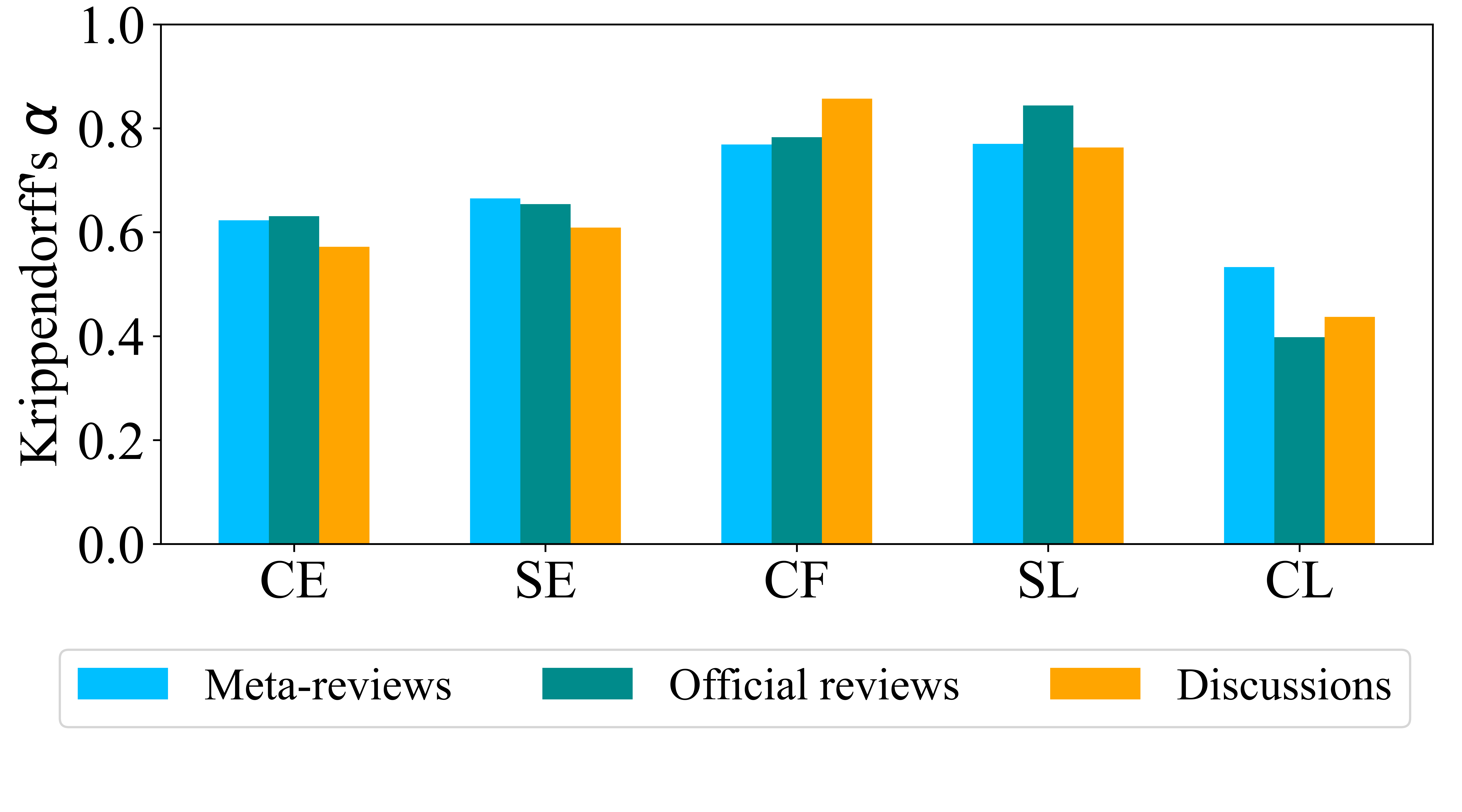}
\caption{Inter-annotator agreement on meta-reviews, official reviews and discussions in terms of Krippendorff's $\alpha$ for different judgement components including Content Expression (CE), Sentiment Expression (SE), Review Facet (RF), Sentiment Level (SL), and Convincingness Level (CL). 
}
\label{fig:iaa}
\end{figure}

In human annotation, there are three types of documents including meta-reviews, official reviews, and discussions (the same definition used in \citet{peersum_2023}) to be annotated. We recruit two annotators\footnote{The two annotators are senior PhD students who are familiar with the peer-review process.} to do this annotation (annotation instructions and design are detailed in Appendix \ref{sec:appendix_annotation_instructions}). 30 samples (i.e.,\ one sample = one meta-review and its corresponding reviews and discussions) are annotated\footnote{Annotating one sample takes about one hour on average and it costs about 60 hours and 2,100 US dollars in total.}, and
in total, we have 1,812 and 1,744 judgements from the two annotators. The statistics of these 30 samples are presented in \tabref{annotation_data_statistics}. We present the agreement of the two annotators in~\figref{iaa}.\footnote{Calculation details and more results in terms of both Cohen's $\kappa$ and Krippendorff's $\alpha$ are in~\tabref{annotation_agreement_metareview},~\tabref{annotation_agreement_official_reviews} and~\tabref{annotation_agreement_discussions} in \appref{sec:appendix_agreement_correlation}.} Generally, we see a moderate to high agreement, suggesting that the annotation task is robust and reproducible.

\begin{figure}[ht]
\centering
\includegraphics[width=0.49\textwidth, trim=0 20 0 0, clip]{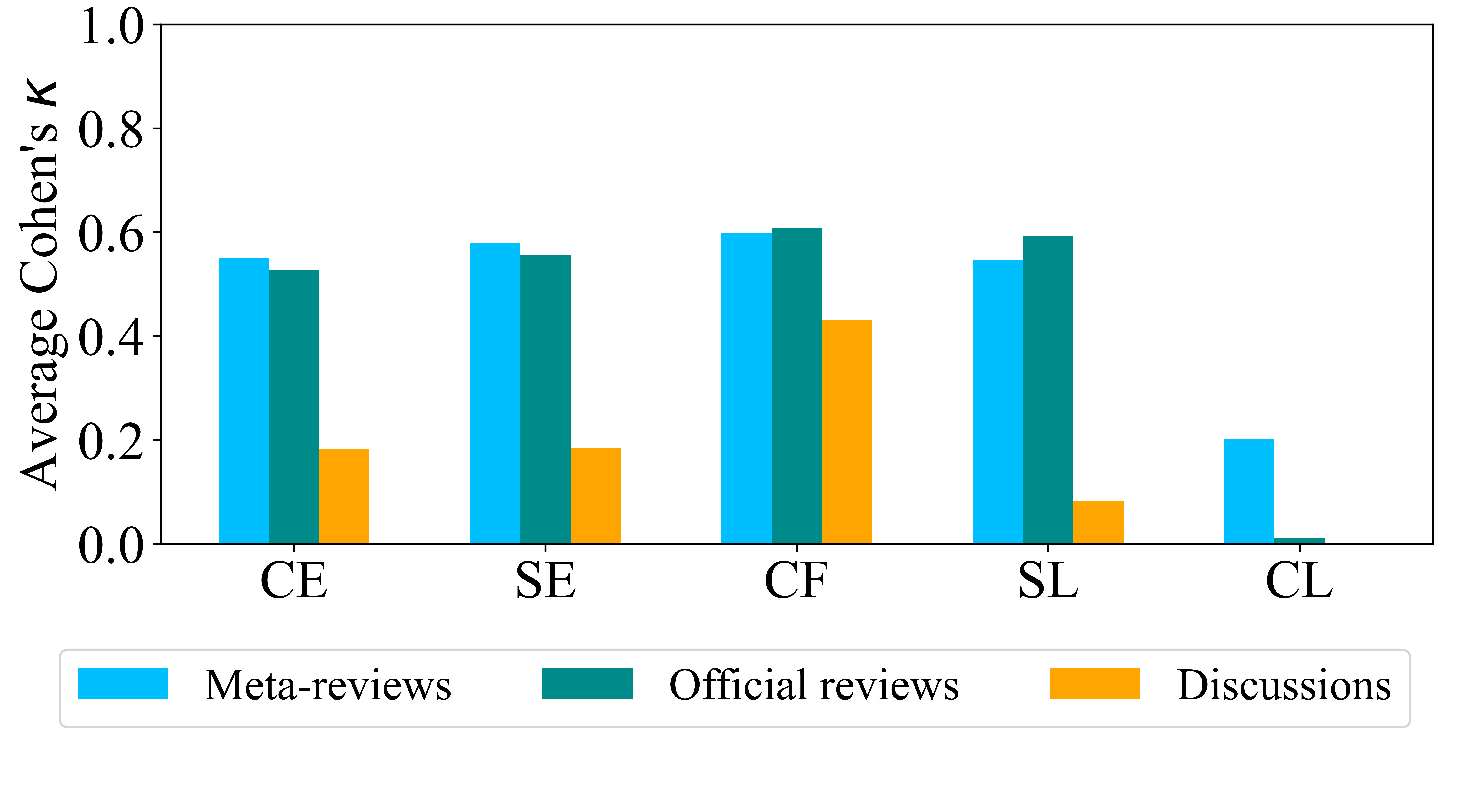}
\caption{The averaged GPT-4's agreement with two human annotators on meta-reviews, official reviews and discussions in terms of Krippendorff's $\alpha$ for different judgement components including Content Expression (CE), Sentiment Expression (SE), Review Facet (RF), Sentiment Level (SL), and Convincingness Level (CL). 
}
\label{fig:gpt4-agreement}
\end{figure}

To get more annotated judgements for further experiments and analysis and investigate whether LLMs can be prompted to identify and extract judgements, 
we split the annotation task into two sub-tasks, extracting content and sentiment expressions and predicting other components of judgements, and use GPT-4~\citep{gpt4_2023}
with in-context learning (see full prompts in \appref{sec:appendix_prompt_expressions} and \ref{sec:appendix_prompt_judgement_predicts} respectively for the two sub-tasks).\footnote{The version of GPT-4 we use is gpt-4-0613.}
We present the average agreement of GPT-4 with the two human annotators in~\figref{gpt4-agreement}.\footnote{More agreement results are in~\tabref{gpt4_agreement_metareviews}, \tabref{gpt4_agreement_official_reviews} and \tabref{gpt4_agreement_discussions} in \appref{sec:appendix_agreement_correlation}.} We can see GPT-4 has a moderate agreement with human annotators for meta-reviews and official reviews, but a low agreement for discussions.
We suspect this may be because the discussions often contain rebuttals which have a different language to reviews or meta-reviews and extracting judgements from them may be more difficult.
Interestingly, we also see that GPT-4 has a poor agreement in terms of convincingness (\figref{gpt4-agreement}), although the human inter-annotator agreement isn't strong in the first place (\figref{iaa}). These observations suggest convincingness is perhaps a subjective assessment.


\subsection{Sentiment Fusion for Consolidation}
\label{sec:sentiment_fusion}

\begin{table}[t]
    \centering
    \begin{adjustbox}{max width=0.9\linewidth}
    \begin{tabular}{lcc}
    \toprule
    \textbf{Facets} & \textbf{\%Judgements} & \textbf{\%Documents} \\
    \midrule
    \textit{Advancement} & 0.2545 & 0.8000\\
    \textit{Soundness} & 0.2786 & 0.7833\\
    \textit{Novelty} & 0.1817 & 0.6833\\
    \textit{Overall} & 0.1414 & 0.5833\\
    \textit{Clarity} & 0.1264& 0.4500\\
    \textit{Compliance} & 0.0174 & 0.0667\\
    \bottomrule
    \end{tabular}
    \end{adjustbox}
    \caption{Frequency of different review facets in meta-review judgements and meta-review documents.}
    \label{tab:dominant_facets_metareview}
\end{table}


With all the annotated judgements extracted by humans and GPT-4, we next dive more into the process of sentiment aggregation. Among all the review facets, we find that \textit{Soundness} and \textit{Advancement} are the two most important review facets when the meta-reviewers write their meta-reviews, while \textit{Compliance} is rarely an issue in meta-reviews (shown in~\tabref{dominant_facets_metareview}). This is consistent with our understanding of the peer-reviewing process.

More importantly, we find that human meta-reviewers do not always follow the majority review sentiment. We find that in PeerSum there are 23.7\% samples where the meta-reviewer's acceptance decision is not consistent with the prediction based on majority voting by review ratings (a sample is defined as consistent when the number of reviews whose rating $\geq 5$ is larger than the number of reviews whose rating $< 5$ and the final decision is \textit{Accept}).
We present an example in \tabref{non_majority_voting_example} where the meta-review does not follow the majority view on \textit{Novelty} from the reviews.

\begin{table}[t]
    \centering
    \small
    \begin{tabular}{|p{7.3cm}|}
    \toprule
    \textbf{\textit{Human-written meta-review sentiment sentence }} \\
    \midrule
    "Although each module in the proposed approach is \underline{\textbf{\textcolor{red}{not novel}}}, it seems that the way they are used to address the specific problem of explainability and especially in text games is \underline{\textbf{\textcolor{ForestGreen}{novel}}} and sound." \\
    \midrule
    \textbf{\textit{All corresponding sentiment texts on Novelty in source reviews and discussions}}\\
    \midrule
    "The generation of temporally extended explanations consists of a cascade of different components, \underline{\textbf{\textcolor{red}{either straightfoward statistics or prior work}}}." \\
    "The novelty is \underline{\textbf{\textcolor{red}{a bit low}}}." \\
    "overall novelty is \underline{\textbf{\textcolor{red}{limitted}}}" \\
    "We contend that all steps are \textbf{\textcolor{ForestGreen}{\uline{individually novel as well as their combination}}}."\\
    "we are \underline{\textbf{\textcolor{ForestGreen}{the first}}} to use knowledge graph attention-based attribution to explain actions in such grounded environments"\\
    \bottomrule
    \end{tabular}
    \caption{The example of a meta-review sentiment on \textit{Novelty} which is not following majority voting of sentiments in source documents. The \underline{\textbf{\textcolor{ForestGreen}{green}}} 
 and \underline{\textbf{\textcolor{red}{red}}} texts indicate positive and negative sentiments, respectively.}
    \label{tab:non_majority_voting_example}
\end{table}


To understand how well the judgements from the source documents (i.e., reviews and discussions) predict the overall sentiments in the meta-reviews for each review facet, we next formulate a text classification task where the output is the sentiment level of a content expression for a review facet in the meta-review, and the input is either: (1) the annotated judgements for the facet from the source documents; or (2) the full text of the source documents. We (zero-shot) prompt GPT-4 (full prompt detailed in \appref{sec:appendix_prompt_sentiment_level_prediction}) with either input to predict 100 randomly sampled human-annotated instances and present the results in \tabref{sentiment_level_prediction_results}. Using judgements only as input, we see that it works better in 4 out of 6 facets --- this preliminary result suggests our framework of extracting these judgements as an intermediate step may help generate better meta-reviews.

\begin{table}[t]
    \centering
    \begin{adjustbox}{max width=0.90\linewidth}
    \begin{tabular}{lcc}
    \toprule
    \textbf{Review Facets} & \textbf{Judgements} & \textbf{Full Texts} \\
    \midrule
    \textit{Advancement} & 0.677 & \textbf{0.697}\\
    \textit{Soundness} & \textbf{0.684} & 0.667\\
    \textit{Novelty} & \textbf{0.700} & 0.650\\
    \textit{Overall} & \textbf{0.643} & 0.631\\
    \textit{Clarity} & \textbf{0.712} & 0.645\\
    \textit{Compliance} & 0.555 & \textbf{0.593}\\
    
    \bottomrule
    \end{tabular}
    \end{adjustbox}
    \caption{Accuracy of GPT-4 in predicting the sentiment levels in meta-reviews for each facet, using either only the annotated judgements (``Judgements'') or the full text (``Full Texts'') from the source documents.}
    \label{tab:sentiment_level_prediction_results}
\end{table}

\section{Sentiment-Aware Evaluation on Information Consolidation}

In this section, we focus more on how to evaluate the sentiments of the generated summaries or meta-reviews in meta-review generation based on our proposed framework. We propose FacetEval and FusionEval which are reference-based and reference-free metrics, respectively.

\subsection{Measuring Sentiment Similarity to Human-Written Meta-Review}
\label{sec:facet_eval}
To assess the quality of generated meta-reviews, we propose a reference-based evaluation metric, FacetEval, measuring the sentiment consistency $c$ between the generated meta-review and the corresponding human-written meta-review in all review facets. Different from the generic evaluation metrics for abstractive summarization or text generation which mostly adopt surface-form matching, we focus more on review facets and their corresponding sentiment levels. 

Specifically, we use the distribution of sentiments in all review facets to represent the meta-review and use the cosine similarity of the two vectors as the final score $s$. 
\begin{align}
    s &= \cos{(\boldsymbol{m}_h, \boldsymbol{m}_g)} \\
    \boldsymbol{m} &= \big\|_{f} [P^+_f, P_f, N^+_f, N_f, O_f]
\end{align}
where $\big\|$ denotes concatenation of representations for different facets, $\boldsymbol{m}_h$ and $\boldsymbol{m}_g$ are representations of the human-written and model-generated meta-reviews respectively. The representation $\boldsymbol{m}$ of the meta-review is the concatenation of vector representations of all review facets. Each facet of the document is represented by the frequency of different sentiment levels on the facet. The facet $f$ is represented by a five-dimension vector $[P^+_f, P_f, N^+_f, N_f, O_f]$ where $P^+_f$ denotes the frequency of \textit{Strongly positive} for the facet $f$, $P_f$ the frequency of \textit{Positive}, $N^+_f$  the frequency of \textit{Strongly negative}, $N_f$  the frequency of \textit{Negative}, and $O_f$  whether this facet is involved in the document. All the sentiments are obtained with GPT-4 following in-context learning in~\secref{judgement_identification_and_extraction}.

Following the similarity of meta-reviews, we could also calculate sentiment consistency among official reviews. Specifically, for every two official reviews $i$ and $j$, the consistency in the facet $f$ is the cosine similarity between two vector representations of documents.
\begin{align}
    c^f_{ij} &= \cos{(\boldsymbol{d}_i, \boldsymbol{d}_j)}
\end{align}
where $\boldsymbol{d}^f = [P^+_f, P_f, N^+_f, N_f, O_f]$. Results shown in~\tabref{sentiment_consistency} suggest that different reviews are consistent in the sentiment to \textit{Compliance} while there is much lower consistency in \textit{Clarity} and \textit{Novelty}. Moreover, we find that conflict reviews\footnote{The same as in PeerSum~\citep{peersum_2023}, if any two reviews have ratings where the gap is larger than 4 they are conflict reviews.} would prefer showing conflicts in \textit{Advancement}, \textit{Novelty}, \textit{Clarity} and \textit{Overall}. This is also consistent with our typical understanding of peer reviews and occasional conflicts among them.

\begin{table}[t]
    \centering
    \begin{adjustbox}{max width=0.95\linewidth}
    \begin{tabular}{lcc}
    \toprule
    \textbf{Review Facet} & \textbf{w/ Conflicts} & \textbf{w/o Conflicts} \\
    \midrule
    \textit{Advancement} & 0.463 (0.135) & 0.551 (0.137)\\
    \textit{Soundness} & 0.526 (0.158) & 0.501 (0.110)\\
    \textit{Novelty} & 0.300 (0.159) & 0.357 (0.168)\\
    \textit{Overall} & 0.433 (0.147) & 0.597 (0.172)\\
    \textit{Clarity} & 0.317 (0.133) & 0.337 (0.145)\\
    \textit{Compliance} & 0.827 (0.071) & 0.771 (0.118)\\
    \bottomrule
    \end{tabular}
    \end{adjustbox}
    \caption{Sentiment consistency among different official reviews. (Variances are in the brackets.)}
    \label{tab:sentiment_consistency}
\end{table}

\begin{table*}[t]
    \setlength\tabcolsep{3pt}
    \centering
    \begin{adjustbox}{max width=0.9\linewidth}
    \begin{tabular}{lcrrrr}
    \toprule
    \textbf{LLM} & \textbf{Evaluation Metric} & \textbf{Prompt-Naive} & \textbf{Prompt-LLM} & \textbf{Prompt-Ours} & \textbf{Pipeline-Ours} \\
    \midrule
    \multirow{5}{*}{GPT-4}
    & FusionEval & 50.14 & 48.90 & \underline{53.62} & \textbf{57.43} \\
     & FacetEval & 35.42 & 40.54 & \underline{41.98} & \textbf{42.36}\\
     & ROUGE-1 & 27.16 & \underline{27.49} & \textbf{28.02} & 24.91\\
     & ROUGE-2 & \textbf{6.63} & 6.03 & \underline{6.57} & 4.57\\
     & ROUGE-L & \underline{24.78} & 24.75 & \textbf{25.51} & 22.70\\
     \midrule
    \multirow{5}{*}{GPT-3.5} 
    & FusionEval & 48.35 & 49.66 & \underline{51.40} & \textbf{55.96} \\
    & FacetEval & 38.44 & 36.83 & \textbf{39.88} & \underline{39.50}\\
     & ROUGE-1 & 28.22 & 25.04 & \textbf{29.56} & \underline{28.92}\\
     & ROUGE-2 & \underline{06.63} & 05.79 & \textbf{6.95} & 5.52\\
     & ROUGE-L & \underline{25.36} & 22.77 & \textbf{26.69} & 16.13\\
     \midrule
    \multirow{3}{*}{LLaMA2-7B} 
    & FusionEval & 46.85 & 46.83 & \underline{50.18} & \textbf{52.68}\\
    & FacetEval & 35.89 & 32.49 & \underline{38.07} & \textbf{38.35}\\
     & ROUGE-1 & \underline{25.94} & 23.88 & \textbf{27.00} & 19.39\\
     & ROUGE-2 & \underline{6.04} & 4.50 & \textbf{6.86} & 4.12\\
     & ROUGE-L & \underline{23.57} & 21.59 & \textbf{24.59} & 17.37\\
     \midrule
    \multirow{3}{*}{LLaMA2-70B} 
    & FusionEval & 47.35 & 48.53 & \underline{50.24} & \textbf{52.80}\\
    & FacetEval & 35.90 & 36.40 & \underline{36.64} & \textbf{36.82}\\
     & ROUGE-1 & \underline{26.61} & 16.60 & \textbf{26.98} & 26.41\\
     & ROUGE-2 & \textbf{6.56} & 3.13 & \underline{5.58} & 4.48\\
     & ROUGE-L & \textbf{24.62} & 14.63 & \underline{24.20} & 23.71\\
     
    \bottomrule
    \end{tabular}
    \end{adjustbox}
    \caption{Performances of different LLMs with different prompting methods. For all metrics, a larger value denotes better performance. The bold and underlined values are the best and second in each row, respectively ($\times 0.01$)}
    \label{tab:automatic_evaluation}
\end{table*}

\begin{table}[t]
    \setlength\tabcolsep{3.5pt}
    \centering
    \begin{adjustbox}{max width=\linewidth}
    \begin{tabular}{lcc}
    \toprule
    \textbf{Competition Groups} & \textbf{Preferred} & \textbf{IAA}\\
    \midrule
    Prompt-Naive LLaMA2-70B & 46.67\% & \multirow{2}{*}{0.64}\\
    Prompt-Ours LLaMA2-70B & 53.33\% \\
    \midrule
    Prompt-Ours GPT-4 & 73.33\% & \multirow{2}{*}{0.74}\\
    Human-Written & 26.67\%\\
    \bottomrule
    \end{tabular}
    \end{adjustbox}
    \caption{Two groups of human evaluation results based on human preferences: (1) comparing generated meta-reviews by Prompt-Naive and Prompt-Ours, and (2) comparing human-written meta-reviews and those generated by Prompt-Ours. IAA denotes inter-annotator agreement calculated with nominal Krippendorff's $\alpha$.}
    \label{tab:human_evaluation}
\end{table}

\begin{table*}[ht]
    \scriptsize
    \centering
    \begin{tabular}{|p{15.5cm}|}
    \toprule
    \textbf{\small Human-Written Meta-Review}\\
    \midrule
    This paper suggests a method for defending against adversarial examples and out-of-distribution samples via projection onto the data manifold. The paper suggests a \underline{\textbf{\small \textcolor{ForestGreen}{new}}} \underline{\textbf{\small \textcolor{blue}{method for detecting when hidden layers are off of the manifold}}}, and uses auto encoders to map them back onto the manifold. \underline{\textbf{\small \textcolor{blue}{The paper}}} is \underline{\textbf{\small \textcolor{ForestGreen}{well-written}}} and \underline{\textbf{\small \textcolor{blue}{the method}}} is \underline{\textbf{\small \textcolor{ForestGreen}{novel and interesting}}}. However, most of the reviewers agree that \underline{\textbf{\small \textcolor{blue}{the original robustness evaluations}}} were \underline{\textbf{\small \textcolor{red}{not sufficient}}} due to restricting the evaluation to using FGSM baseline and comparison with thermometer encoding (which both are known to not be fully effective baselines). After rebuttal, Reviewer 4 points out that \underline{\textbf{\small \textcolor{red}{the method offers very little}}} \underline{\textbf{\small \textcolor{blue}{robustness over adversarial training alone}}}, even though it is combined with adversarial training, which suggests that \underline{\textbf{\small \textcolor{blue}{the method itself}}} \underline{\textbf{\small \textcolor{red}{provides very little robustness}}}. \\
    \midrule
    \textbf{\small Meta-Review Generated by GPT-4 with Prompt-Naive}  \\
    \midrule
    The reviewers have raised \underline{\textbf{\small \textcolor{red}{several concerns}}} \textbf{\small \textcolor{blue}{\uline{about the empirical evaluation and effectiveness of the proposed fortified network model}}}. They note that \underline{\textbf{\small \textcolor{blue}{the benefits of the fortified layers}}} seem to be \underline{\textbf{\small \textcolor{red}{marginal}}} and \underline{\textbf{\small \textcolor{blue}{the experimental evaluation}}} \textbf{\small \textcolor{red}{\uline{doesn't sufficiently demonstrate that the increased robustness is due to the fortified layers rather than the adversarial training}}}. Furthermore, most of the reviewers \textbf{\small \textcolor{red}{\uline{would like to see}}} \textbf{\small \textcolor{blue}{\uline{benchmarks against state-of-the-art defenses such as PGD}}}. Among other issues, \textbf{\small \textcolor{red}{\uline{reviewers have questioned}}} \textbf{\small \textcolor{blue}{\uline{the use of an autoencoder and the decision to use a denoising module in the hidden layers}}}. Simply put, \textbf{\small \textcolor{red}{\uline{the reviewers would like a stronger reasoning for}}} \textbf{\small \textcolor{blue}{\uline{the methodological choices}}}. There are also \textbf{\small \textcolor{red}{\uline{requests for clarification}}} on \textbf{\small \textcolor{blue}{\uline{the denoising terms and a comparison with a simple regularization loss}}}. Moreover, a reviewer also \textbf{\small \textcolor{red}{\uline{raised a question}}} on \textbf{\small \textcolor{blue}{\uline{the effectiveness of the defense method}}} against other types of attacks that might potentially compromise the proposed network. Most reviewers felt that \textbf{\small \textcolor{blue}{\uline{the paper}}} \textbf{\small \textcolor{red}{\uline{does not provide enough evidence that defends the proposed fortified network against adversarial attacks}}}. Finally, \textbf{\small \textcolor{red}{\uline{concerns}}} about \textbf{\small \textcolor{blue}{\uline{the presentation of results and tables}}}, and \textbf{\small \textcolor{blue}{\uline{grammar errors}}} have also \textbf{\small \textcolor{red}{\uline{been pointed out}}}. In conclusion, while the reviewers \textbf{\small \textcolor{ForestGreen}{\uline{acknowledge}}} \textbf{\small \textcolor{blue}{\uline{the relevance and importance of the research problem}}}, \textbf{\small \textcolor{blue}{\uline{the manuscript}}} \textbf{\small \textcolor{red}{\uline{requires significant amendments}}} to convince the experts about the effectiveness of the proposed method in tackling the problem. The authors must address these concerns and improve the manuscript to make a compelling case for their proposed approach. \\
    \midrule
    \textbf{\small Meta-Review Generated by GPT-4 with Prompt-Ours}\\
    \midrule
    The paper at hand proposes a fortified network model as a defense against adversarial attacks, based on the concept of denoising autoencoders. A \textbf{\small \textcolor{ForestGreen}{\uline{key novelty}}} in the presented approach is \textbf{\small \textcolor{blue}{\uline{the placement of the denoising feature within the hidden layers}}}, rather than just the input layer.The \textbf{\small \textcolor{blue}{\uline{clarity and structure}}} of the document have been \textbf{\small \textcolor{ForestGreen}{\uline{appreciated}}} by a number of reviewers, particularly the experimental data presentation and robust bibliography. There are, however, a few areas \textbf{\small \textcolor{red}{\uline{noted for improvement}}} - predominantly around \textbf{\small \textcolor{blue}{\uline{grammatical mistakes and confusing sentence structure}}}. A significant source of debate among reviewers centers on the empirically evaluated efficacy of the fortified model. The authors propose that their model improves robustness by up to 5\% over the baseline, but several reviewers argue that \textbf{\small \textcolor{blue}{\uline{this improvement}}} is \textbf{\small \textcolor{red}{\uline{marginal and does not sufficiently justify the implementation of an additional objective}}}. This argument is most prominent when looking at the study's use of non-standard models and a PGD baseline that performs below the state-of-the-art, leading to questions about the accuracy of the baseline evaluation. Further, accuracy results against an Linf adversary of eps>0.1 on CIFAR-10 raises \textbf{\small \textcolor{red}{\uline{questions}}} around \textbf{\small \textcolor{blue}{\uline{the experiment's setup}}} given known limitations relating to image perturbations. \textbf{\small \textcolor{red}{\uline{Suggestions were made}}} to \textbf{\small \textcolor{blue}{\uline{explore a nearest neighbor attack from the test set, multiple gradient queries per PGD step, and report standard non-robust accuracies}}} to ensure robustness was not an artefact of model expressivity.Lastly, some apprehension was expressed about the motivation behind and efficacy of using a denoising autoencoder, \textbf{\small \textcolor{red}{\uline{with calls for}}} \textbf{\small \textcolor{blue}{\uline{comparison to a simpler baseline to justify the approach}}}. In conclusion, while the fortified network model presents an \textbf{\small \textcolor{ForestGreen}{\uline{intriguing}}} \textbf{\small \textcolor{blue}{\uline{approach}}} to defending against adversarial attacks, the reviewers \textbf{\small \textcolor{red}{\uline{do not overwhelmingly agree}}} that \textbf{\small \textcolor{blue}{\uline{the empirical results sufficiently demonstrate advancement over existing methods}}}. The majority find the defensive gain too marginal given the additional complexity and question the setup of the empirical evaluation. Further clarity in method and expanded empirical evidence would facilitating a stronger case for the proposed model. \\
    \bottomrule
    \end{tabular}
    \caption{Human-written meta-review and the corresponding meta-reviews generated by GPT-4 with the naive prompt and the prompt based on the sentiment consolidation. (The \underline{\textbf{\textcolor{ForestGreen}{green spans}}} are positive sentiment values, \underline{\textbf{\textcolor{red}{red spans}}} are negative sentiment values, while \underline{\textbf{\textcolor{blue}{blue spans}}} are the content expressions.)}
    \label{tab:case_study}
\end{table*}

\subsection{Measuring Sentiment Fusion for Individual Facets} 
\label{sec:fusion_eval}

Sentiments in the generated meta-reviews should be in line with the aggregate sentiment from the individual source documents including reviews and discussions. Seeing GPT-4 can predict the overall sentiment using judgements from source documents (Section~\ref{sec:sentiment_fusion}), we introduce a reference-free evaluation metric, FusionEval, which assesses the consistency between the sentiments in the generated meta-review and that predicted by GPT-4 (with zero-shot prompting) from the source documents. Higher consistency implies the overall sentiment in generated meta-reviews are representative of the sentiments in the reviews and discussions (source documents).

Specifically, we first extract judgements from the generated meta-review following~\secref{judgement_identification_and_extraction}, and these judgements consist of \textit{Content Expressions}, $E$ and \textit{Sentiment Levels}, $L$, and the corresponding \textit{Review Facets}, $F$. Next, for each expression, $e\in E$, we predict the \textit{Sentiment Level}, $l'$, using GPT-4 (zero-shot) based on all judgements for the same \textit{Review Facet} in the source documents following~\secref{sentiment_fusion}, and we get predicted \textit{Sentiment Levels} for all judgements, $L'$.
Lastly, FusionEval computes an accuracy score by evaluating $L'$ against $L$. FusionEval only considers the precision instead of the recall for meta-review sentiments as it is reference-free and we have no information about the count of judgements that should be synthesized.

\section{Enhancing LLMs with Explicit Information Consolidation}

In this section, we propose two prompting methods to integrate the sentiment consolidation framework to generate meta-reviews. We compare the two methods with other prompting strategies including naive prompting and prompting with LLM-generated logic. We also run experiments on open-source models besides OpenAI ones to investigate the influence of different prompting methods on different models. The experiments are based on automatic and human annotation on 500 samples from PeerSum.\footnote{To avoid data contamination, we only use samples which were produced in and after 2022.}

\subsection{Prompting LLMs with Sentiment Consolidation Logic}


Following the process in \figref{three-layer-framework} we propose decomposing the meta-review generation process in the following steps: (1) Extracting content and sentiment expressions of judgements from source documents; (2) Predicting \textit{Review Facets}, \textit{Sentiment Levels}, and \textit{Convincingness Levels}; (3) Clustering extracted judgements for different review facets; (4) Generating a ``mini summary'' for judgements on the same review facet; and (5) Generating the final meta-review based on the mini summaries for all review facets.

We explore two methods to integrate this process for prompting an LLM. (1) Prompt-Ours: we describe the five steps in a single prompt and ask GPT-4 to generate the final meta-reviews (full prompt in \appref{prompt_logic}); (2) Pipeline-Ours: we create one prompt for each of the five steps, where the input for the intermediate step is the output from the previous step (full prompts in \appref{prompt_pipeline}). 

We experiment with four open-source and close-source LLMs: GPT-4, GPT-3.5, LLaMA2-70B and LLaMA2-7B.\footnote{Precise model names for them are: gpt-4-0613, gpt-3.5-turbo-1106, LLaMA2-70B-Chat, LLaMA2-7B-Chat. Note that for Pipeline-Ours, we always use GPT-4 for the first two steps, as we find that the other LLMs perform poorly for these tasks.}

\subsection{Baselines}

As baselines, we include two more methods: (1) Prompt-Naive: which prompts an LLM with a simple instruction to generate the meta-review (full prompt in~\appref{prompt_naive}); and (2) Prompt-LLM: where we ask an LLM to self-generate the detailed steps for meta-review generation and we include these steps in the final prompt for meta-review generation (full prompt in~\appref{prompt_llm}).

\subsection{Reference-Based and Reference-Free Automatic Evaluation}

For automatic evaluation, we adopt ROUGE~\citep{rouge_2003}\footnote{We use the implementation of the algorithm in \url{https://pypi.org/project/rouge-score/}}, FacetEval (\secref{facet_eval}) and FusionEval (\secref{fusion_eval}).\footnote{We do not consider other metrics such as BERTScore~\citep{bertscore_2020}, UniEval~\citep{unieval_2022}, or G-Eval~\citep{geval_2021} as they have only been validated for summarization in news or a more general domain.}

We present the results in~\tabref{automatic_evaluation}. Most LLMs perform better with either of our prompting methods (Prompt-Ours and Pipeline-Ours) than the baselines (Prompt-Naive and Prompt-LLM). Comparing between Prompt-Ours and Pipeline-Ours, the former seems to do best for ROGUE while the latter for FusionEval/FacetEval. As FusionEval/FacetEval assesses the accuracy of the overall sentiment specifically, Pipeline-Ours is marginally better when it comes to getting the overall sentiment correct.
Comparing different LLMs, GPT-4 seems to work best, although that is mostly true for FusionEval/FacetEval.


\subsection{Reference-Free Human Evaluation}
To further validate the effectiveness of our prompting methods, we conduct human evaluations to assess the quality of meta-reviews generated by different prompting methods or written by human meta-reviewers. We recruited three volunteer annotators who are senior PhD students familiar with artificial intelligence research and the peer review process. They are asked to select their preferred meta-reviews based on their own understanding of high-quality meta-reviews without knowing the source.\footnote{We use majority voting to get the final human preference.}

\paragraph{Prompt-Naive vs Prompt-Ours} We randomly select 30 samples and the annotators are asked to compare the generated meta-reviews by Prompt-Naive and Prompt-Ours (using LLaMA2-70B) and select which one is better. \tabref{human_evaluation} shows that the meta-reviews generated by Prompt-Ours are selected more by the annotators.

\paragraph{Prompt-Ours vs Human-Written} We repeat the same experiments, but this time comparing meta-reviews generated by Prompt-Ours (GPT-4) vs.\ written by humans. Looking at \tabref{human_evaluation}, interestingly Prompt-Ours are much more preferred by the annotators. We suspect this may be because the generated meta-reviews tend to be more consistent in terms of the amount of detail it writes for each review facet, where else there is more variance for the human-written meta-reviews.

\subsection{Case Study on Generated Meta-Reviews}

To dive deeper into what difference the integration of sentiment consolidation framework makes, we also conduct a case study on generated meta-reviews with different prompting methods. We find that generated meta-reviews all seem plausible and machine-generated meta-reviews are much longer than human-written ones.  In machine-generated meta-reviews, there are more details which are sometimes unnecessary or redundant. As shown in the example in~\tabref{case_study}, details such as "PGD" or "CIFA-10" are not essential to form the meta-review. 

Our proposed Prompt-Ours tend to have a more balanced judgements. For example, in~\tabref{case_study}, Prompt-Naive does not talk about the  positive aspects for \textit{Clarity} and only highlights some issues, but
Prompt-Ours comments on both the strengths and weaknesses for \textit{Clarity}.
This is consistent with the finding in~\tabref{automatic_evaluation} that Prompt-Naive gets worse sentiments than Prompt-Ours.

\section{Conclusions and Future Work}
In this paper, we explore sentiment-focused multi-document information consolidation within the task of scientific sentiment summarization. We introduce a three-layer framework of sentiment consolidation to focus on generating meta-reviews and it considers the sentiments for each review facet in the reviews and discussions. We also propose automatic evaluation metrics that assess the overall sentiments in the generated meta-reviews. Experiments on meta-review generation show that prompting LLMs by following the processes in the three-layer framework results in better meta-reviews, providing an empirical validation of our framework for describing the meta-review writing process. As the sentiment consolidation also exist in other domains where human reviews or comments exist such as politics and advertisement, we will explore adapting our proposed sentiment consolidation framework into other domains in the future.

\section*{Limitations}
Although integration of the sentiment consolidation framework could improve the generation results, there are still some limitations of this work.
\squishlist
    \item As in other areas peer review data is not publicly available, we use the data only from some artificial intelligence conferences, and this may make the models biased. We hope that more data from diverse areas could be included.
    \item Experiments are only in English texts rather than other languages.
    \item We only inject the information consolidation logic into prompting based models instead of fine-tuning based models. We will investigate leveraging the information consolidation framework to improve fine-tuned models in the future.
    \item Although GPT-4 can predict meta-review sentiments based on source judgements to some extent, we have to understand more about how these models achieve this and what makes them fail in error cases.
    \item Meta-review generation is not only about sentiment prediction, future work has to consider more information such as argumentation in source reviews and justification in meta-reviews.
\squishend

\section*{Ethics Statement}

While our experiments demonstrate that the models exhibit potential in generating satisfactory meta-reviews to a certain degree, we strongly advise against solely relying on the generated results without manual verification and review, as instances of hallucinations exist in the generations. It is important to emphasize that we do not advocate for replacing human meta-reviewers with LLMs. However, it is noteworthy that these models have the capacity to enhance the meta-reviewing process, rendering it more efficient and effective.

\section*{Acknowledgements}

We would thank all the anonymous reviewers for their great comments to improve the work. It is impossible to have this work without the volunteer contributions of the annotators on human evaluations.


\bibliography{custom, mreferences}
\bibliographystyle{acl_natbib}

\newpage

\onecolumn
\appendix

\section{Review Criteria in Different Reviewer Guidelines}
\label{appendix_review_guidelines}

\begin{table}[h]
    \centering
    \begin{adjustbox}{max width=\linewidth}
    \begin{tabular}{p{50mm}p{120mm}<{\centering}}
    \toprule
    \textbf{Academic Press} & \textbf{Review guidelines}\\
    \midrule
    ACM & \url{https://dl.acm.org/journal/dgov/reviewer-guidelines } \\
    ACL Rolling Review & \url{https://aclrollingreview.org/reviewertutorial} \\
    IEEE & \url{https://conferences.ieeeauthorcenter.ieee.org/understand-peer-review/} \\
    Springer & \url{https://www.springer.com/gp/authors-editors/authorandreviewertutorials/howtopeerreview/evaluating-manuscripts/10286398} \\
    NeurIPS & \url{https://neurips.cc/Conferences/2021/Reviewer-Guidelines} \\
    ICLR & \url{https://iclr.cc/Conferences/2023/ReviewerGuide#Reviewing instructions} \\
    ACL & \url{https://2023.aclweb.org/blog/review-acl23/ } \\
    Cambridge University Press & \url{https://www.cambridge.org/core/services/aop-file-manager/file/5a1eb62e67f405260662a0df/Refreshed-Guide-Peer-Review-Journal.pdf}\\
    \bottomrule
    \end{tabular}
    \end{adjustbox}
    \caption{Review guidelines from different academic presses.}
    \label{tab:review_guidelines}
\end{table}

\section{Annotation Instructions for Human Annotation}
\label{sec:appendix_annotation_instructions}

The screenshots of the two-page annotation instruction for human annotation are shown in \figref{annotation-instruction-p1} and \figref{annotation-instruction-p2} in the last two pages of the Appendix.

\section{Inter-Annotator Agreement Among Human Annotators and GPT-4}
\label{sec:appendix_agreement_correlation}

We describe how we calculate inter-annotator agreement among human annotators and GPT-4 here. For Content Expression and Sentiment Expression, as they are highlighted text spans we calculate the character-level agreement with Krippendorf's $\alpha$ and Cohen's $\kappa$. Specifically, for each document, two annotators may highlight different text spans for Content Expression and Sentiment Expression. We construct two vectors of the same length as the characters to represent the highlighting behaviours of any two annotators. This agreement shows whether annotators identify sentiments from similar text spans. 

For \textit{Review Facet}, \textit{Sentiment Level}, and \textit{Convincingness Level}, we calculate Krippendorf's $\alpha$ and Cohen's $\kappa$ in a common way. We first identify whether two annotators recognize sentiment from the same text span with a ROUGE threshold (the summation of ROUGE-1, ROUGE-2 and ROUGE-L between highlighted text spans for sentiment is larger than 2.0), and calculate agreement on the predicted values.

Inter-annotator agreement between two human annotators for human annotation in Section~\ref{sec:judgement_identification_and_extraction} are present in~\tabref{annotation_agreement_metareview}, ~\tabref{annotation_agreement_official_reviews}, and \tabref{annotation_agreement_discussions}. Averaged agreement of GPT-4 with the two human annotators are present in~\tabref{gpt4_agreement_metareviews}, ~\tabref{gpt4_agreement_official_reviews}, and~\tabref{gpt4_agreement_discussions}.

\begin{table}[]
    \centering
    \begin{adjustbox}{max width=\linewidth}
    \begin{tabular}{lcc}
    \toprule
    \textbf{Annotation} & \textbf{Cohen's $\kappa$} & \textbf{Krippendorf's $\alpha$} \\
    \midrule
    Content Expression & 0.623 & 0.623 \\
    Sentiment Expression & 0.666 & 0.665 \\
    Review Facet & 0.769 & 0.769 \\
    Sentiment Level & 0.770 & 0.770 \\
    Convincingness Level & 0.534 & 0.533 \\
    \bottomrule
    \end{tabular}
    \end{adjustbox}
    \caption{Human annotator agreement on annotating meta-reviews.}
    \label{tab:annotation_agreement_metareview}
\end{table}

\begin{table}[]
    \centering
    \begin{adjustbox}{max width=\linewidth}
    \begin{tabular}{lcc}
    \toprule
    \textbf{Annotation} & \textbf{Cohen's $\kappa$} & \textbf{Krippendorff's $\alpha$} \\
    \midrule
    Content Expression & 0.631 & 0.631 \\
    Sentiment Expression & 0.654 & 0.654 \\
    Review Facet & 0.783 & 0.783 \\
    Sentiment Level & 0.844 & 0.844 \\
    Convincingness Level & 0.405 & 0.398\\
    \bottomrule
    \end{tabular}
    \end{adjustbox}
    \caption{Human annotator agreement on annotating official reviews.}
    \label{tab:annotation_agreement_official_reviews}
\end{table}

\begin{table}[]
    \centering
    \begin{adjustbox}{max width=\linewidth}
    \begin{tabular}{lcc}
    \toprule
    \textbf{Annotation} & \textbf{Cohen's $\kappa$} & \textbf{Krippendorff's $\alpha$} \\
    \midrule
    Content Expression & 0.572 & 0.572\\
    Sentiment Expression & 0.609 & 0.609\\
    Review Facets & 0.857 & 0.857\\
    Sentiment Levels & 0.764 & 0.763\\
    Convincingness Levels & 0.455& 0.437\\
    \bottomrule
    \end{tabular}
    \end{adjustbox}
    \caption{Human annotator agreement on annotating discussions.}
    \label{tab:annotation_agreement_discussions}
\end{table}

\begin{table}[]
    \centering
    \begin{adjustbox}{max width=\linewidth}
    \begin{tabular}{lccc}
    \toprule
    \textbf{Annotation} & \textit{A} & \textit{B} & \textbf{Avg} \\
    \midrule
    Content Expression & 0.558 & 0.542 & 0.550 \\
    Sentiment Expression & 0.565 & 0.594 & 0.580 \\
    Review Facets & 0.588 & 0.610 & 0.599 \\
    Sentiment Levels & 0.552 & 0.541 & 0.547\\
    Convincingness Levels & 0.213 & 0.192 & 0.203\\
    \bottomrule
    \end{tabular}
    \end{adjustbox}
    \caption{GPT-4 agreement in terms of Cohen's $\kappa$ with human annotators \textit{A} and \textit{B} on annotating meta-reviews.}
    \label{tab:gpt4_agreement_metareviews}
\end{table}

\begin{table}[]
    \centering
    \begin{adjustbox}{max width=\linewidth}
    \begin{tabular}{lccc}
    \toprule
    \textbf{Annotation} & \textit{A} & \textit{B} & \textbf{Avg} \\
    \midrule
    Content Expression & 0.522 & 0.534 & 0.528\\
    Sentiment Expression & 0.544 & 0.569 & 0.557\\
    Review Facets & 0.579 & 0.637 & 0.608\\
    Sentiment Levels & 0.594 & 0.589 & 0.592\\
    Convincingness Levels & 0.008 & 0.013 & 0.011\\
    \bottomrule
    \end{tabular}
    \end{adjustbox}
    \caption{GPT-4 agreement in terms of Cohen's $\kappa$ with human annotators \textit{A} and \textit{B} on annotating official reviews.}
    \label{tab:gpt4_agreement_official_reviews}
\end{table}

\begin{table}[]
    \centering
    \begin{adjustbox}{max width=\linewidth}
    \begin{tabular}{lccc}
    \toprule
    \textbf{Annotation} & \textit{A} & \textit{B} & \textbf{Avg} \\
    \midrule
    Content Expression & 0.176 & 0.187 & 0.182\\
    Sentiment Expression & 0.182 & 0.188 & 0.185\\
    Review Facets & 0.480 & 0.381 & 0.431\\
    Sentiment Levels & 0.123 & 0.046 & 0.082\\
    Convincingness Levels & 0.0 & 0.0 & 0.0\\
    \bottomrule
    \end{tabular}
    \end{adjustbox}
    \caption{GPT-4 agreement in terms of Cohen's $\kappa$ with human annotators \textit{A} and \textit{B} on annotating discussions.}
    \label{tab:gpt4_agreement_discussions}
\end{table}

\begin{figure}[ht]
\centering
\includegraphics[width=1.0\textwidth, trim=70 60 60 70, clip]{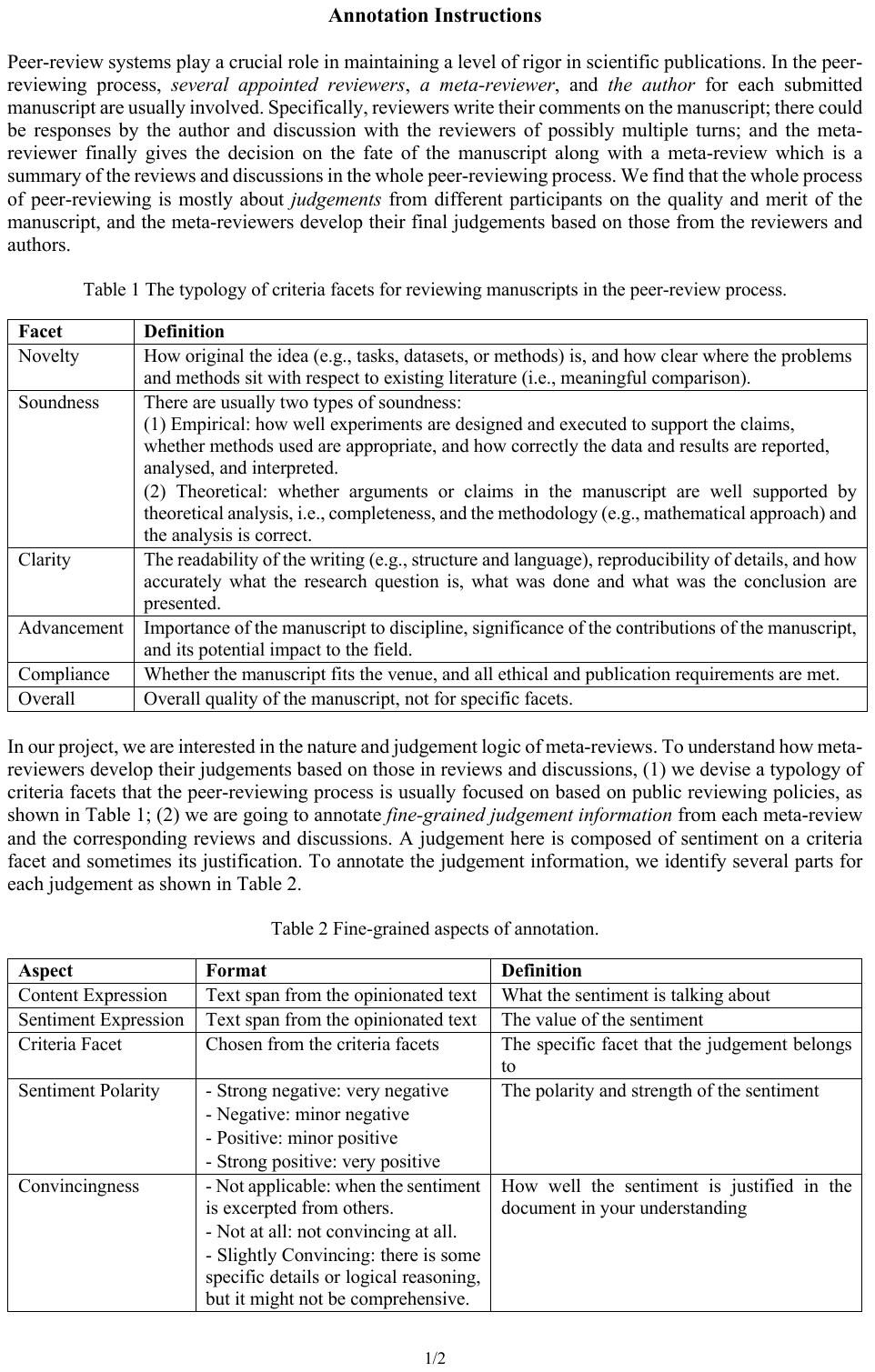}
\caption{The first page of the annotation instruction for human judgement annotation. 
}
\label{fig:annotation-instruction-p1}
\end{figure}

\begin{figure}[ht]
\centering
\includegraphics[width=1.0\textwidth, trim=70 200 60 70, clip]{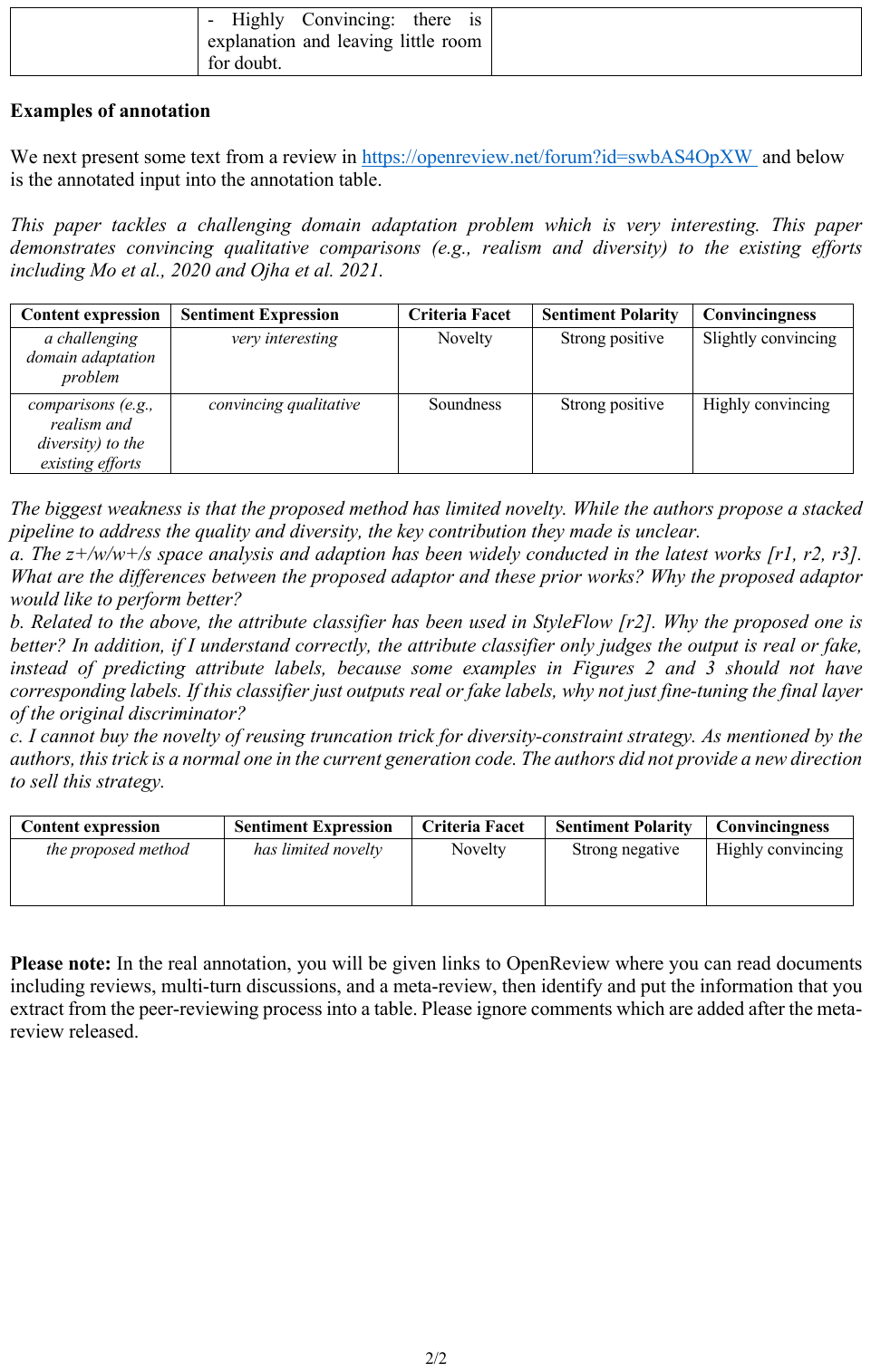}
\caption{The first page of the annotation instruction for human judgement annotation. 
}
\label{fig:annotation-instruction-p2}
\end{figure}

\section{Prompt to Get Content and Sentiment Expressions with GPT-4}
\label{sec:appendix_prompt_expressions}
\begin{lstlisting}[breaklines=true, frame=shadowbox, numbers=left, numberstyle=\small]
Please read the document:

{{source_document}}

This task requires you to analyze the above document which is used to express opinions on the quality of a scientific manuscript. You are good at understanding the sentiment information with judgements in the document.
Please first identify the sentence with judgements only on the quality of scientific manuscripts based on the review facets for scientific peer-review: novelty, soundness, clarity, advancement, compliance and overall quality within the given document.
Once you have found a sentence that provides judgement in one or more of these areas, you then need to extract the specific expression of sentiment and the content it refers to.

The process can be broken into two steps:
1) Identify a judgement sentence that focuses on the quality of the manuscript based on the given criteria.

2) From the identified judgement sentence, extract two pieces of information: the sentiment expression and the content expression. The sentiment expression is the specific term or phrase that conveys the sentiment or opinion. The content expression pertains to the content that this sentiment is referring to.

Please provide the data in the following format:
{"judgement_sentence": "sentence", "content_expression": "content", "sentiment_expression": "sentiment"}

Here are a few examples for your reference:
{"judgement_sentence": "The writing of the paper is not well-written.", "content_expression": "The writing of the paper", "sentiment_expression": "not well-written"}
{"judgement_sentence": "Experimental results are not sufficiently substantiated.", "content_expression": "Experimental results", "sentiment_expression": "not sufficiently substantiated"}
{"judgement_sentence": "This paper presents two novel approaches to provide explanations for the similarity between two samples based on 1) the importance measure of individual features and 2) some of the other pairs of examples used as analogies.", "content_expression": "approaches", "sentiment_expression": "novel"}

The predicted judgments (following the same jsonline format of the above example):
\end{lstlisting}

\section{Prompt to Get Judgement Component Predictions with GPT-4}
\label{sec:appendix_prompt_judgement_predicts}
\begin{lstlisting}[breaklines=true, frame=shadowbox, numbers=left, numberstyle=\small]
Please first read the document below:

{{source_document}}


Please predict the facet that the given judgements are talking about. You can refer to the context in the above source document.

Possible facets:

Novelty: How original the idea (e.g., tasks, datasets, or methods) is, and how clear where the problems and methods sit with respect to existing literature (i.e., meaningful comparison).

Soundness: (1) Empirical: how well experiments are designed and executed to support the claims, whether methods used are appropriate, and how correctly the data and results are reported, analysed, and interpreted. (2) Theoretical: whether arguments or claims in the manuscript are well supported by theoretical analysis, i.e., completeness and the methodology (e.g., mathematical approach) and the analysis is correct.

Clarity: The readability of the writing (e.g., structure and language), reproducibility of details, and how accurately what the research question is, what was done and what was the conclusion are presented.

Advancement: Importance of the manuscript to discipline, significance of the contributions of the manuscript, and its potential impact to the field.

Compliance: Whether the manuscript fits the venue, and all ethical and publication requirements are met.

Overall: Overall quality of the manuscript, not for specific facets.


You are also good at understanding sentiment information in the judgements.

Please predict the original expresser of the sentiment in the judgement sentence. You can refer to the context in the source document.

Possible sentiment expressers:

- Self: the sentiment is from the speaker
- Others: the sentiment is quoted from others


Please predict how well the sentiment in the judgement sentence is justified in the document in your understanding. You can refer to the context in the source document.

Possible sentiment convincingness:

- Not applicable: the sentiment is explicitly excerpted from others.
- Not at all: not convincing at all or when there is no justification. How well the sentiment is justified in the document in your understanding
- Slightly Convincing: there is some evidence or logical reasoning, but it might not be comprehensive.
- Highly Convincing: leaving little room for doubt.


Please predict the polarity and strength of the sentiment in the judgement sentence. You can refer to the context in the source document.

Possible sentiments polarities:

- Strong negative: very negative
- Negative: minor negative
- Positive: minor positive
- Strong positive: very positive


Judgements:
{{judgement_expressions}}

Your predictions for the above judgements (following the same jsonlines format, return the same number of lines, and keep the same content and sentiment expressions):
\end{lstlisting}

\section{Prompts to Predict Meta-Review Sentiment Levels}
\label{sec:appendix_prompt_sentiment_level_prediction}

\subsection{Prediction with Judgements of Source documents}
The judgements are extracted from source documents, and they are in the same review facet to the target meta-review judgement.

\begin{lstlisting}[breaklines=true, frame=shadowbox, numbers=left, numberstyle=\small]
You will be given source judgements from reviewers for a scientific manuscript. Your task is to implicitly write a meta-review for these judgements and predict the sentiment level based on these judgements.

Source Judgements:

{{source_judgements}}

Candidate Sentiment Levels:

- Strong negative
- Negative
- Positive
- Strong positive

Content Expression:

{{content_expression}}

Predict the sentiment level of the given content expression based on the above judgements. You must follow the following format.
{"Content Expression": the above content expression, "Sentiment Level": your predicted sentiment level}
\end{lstlisting}

\subsection{Prediction with Full Texts of Source documents}
The source texts are the concatenation of the source documents.

\begin{lstlisting}[breaklines=true, frame=shadowbox, numbers=left, numberstyle=\small]
You will be given multiple review documents for a scientific manuscript. Your task is to implicitly write a meta-review and  predict the sentiment level based on these documents.

Source Documents:

{{source_texts}}

Candidate Sentiment Levels:

- Strong negative
- Negative
- Positive
- Strong positive

Content Expression:

{{content_expression}}

Predict the sentiment level of the given content expression based on related information in the above documents. You must follow the following format.
{"Content Expression": the above content expression, "Sentiment Level": your predicted sentiment level}
\end{lstlisting}

\section{Prompts for Meta-Review Generation with Integration of Information Consolidation Logic}

\subsection{Prompt with Descriptive Consolidation Logic}
\label{prompt_logic}

\begin{lstlisting}[breaklines=true, frame=shadowbox, numbers=left, numberstyle=\small]
    Your task is to write a meta-review based on the following reviews and discussions for a scientific manuscript.

{{input_documents}}

Following the underlying steps below will get you better generated meta-reviews.

1. Extracting content and sentiment expressions of judgements in all above review and discussion documents;

2. Predicting Review Facets, Sentiment Levels, and Convincingness Levels;
Candidate review facets: Novelty, Soundness, Clarity, Advancement, Compliance, and Overall quality
Candidate sentiment levels: Strong negative, Negative, Positive and Strong positive
Candidate convincingness levels:  Not at all, Slightly Convincing, Highly Convincing

3. Reorganize extracted judgements in different clusters for different review facets;

4. Generate a small summary for judgements on the same review facet with comparison and aggregation;

5. Aggregate judgements in different review facets and write a meta-review based on the aggregation.


You may follow these steps implicitly and only need to output the final meta-review. The final meta-review:
\end{lstlisting}

\subsection{Prompts Used in the Pipeline Generation}
\label{prompt_pipeline}

Prompts for the first two steps, getting content and sentiment expressions and predicting other judgement components, are the same as prompts in Appendix~\ref{sec:appendix_prompt_expressions} and Appendix~\ref{sec:appendix_prompt_judgement_predicts}, respectively.

For the step of generating sub-summaries for individual facets, the prompt is as follows.

\begin{lstlisting}[breaklines=true, frame=shadowbox, numbers=left, numberstyle=\small]
{{input_judgements}}

Write a summary of the above judgements on {{criteria_facet}} of a manuscript.
\end{lstlisting}

For the step of generating final meta-reviews based on sub-summaries of individual facets, the prompt is as follows.

\begin{lstlisting}[breaklines=true, frame=shadowbox, numbers=left, numberstyle=\small]
{{input_sub_summaries}}

Write a meta-review to summarize the above sub-summaries of reviews and discussions in different review facets for a manuscript.
\end{lstlisting}

\subsection{Prompts from Prompt-Naive}
\label{prompt_naive}

For Prompt-Naive in our experiments, the prompt we use is as follows.

\begin{lstlisting}[breaklines=true, frame=shadowbox, numbers=left, numberstyle=\small]
{{input_documents}}

Write a meta-review based on the above reviews and discussions for a manuscript.
\end{lstlisting}

\subsection{Prompts from Prompt-LLM}
\label{prompt_llm}

For Prompt-LLM, we have to generate first the steps with GPT-4 and then the meta-review based on the generated steps.

The prompt to generate the steps: 
\begin{lstlisting}[breaklines=true, frame=shadowbox, numbers=left, numberstyle=\small]
{{input_documents}}

What are the steps to write a meta-review specifically for the above reviews and discussions of a manuscript.
\end{lstlisting}

The prompt to generate the meta-review:

\begin{lstlisting}[breaklines=true, frame=shadowbox, numbers=left, numberstyle=\small]
{{input_documents}}

Follow the following steps and write a meta-review based on the above reviews and discussions for a manuscript.

{{generated_steps}}
\end{lstlisting}

\end{document}